\documentclass[conference]{IEEEtran}
\IEEEoverridecommandlockouts
\usepackage[T1]{fontenc}   
\usepackage[utf8]{inputenc} 
\usepackage{verbatim}
\usepackage{cite}
\usepackage{makecell}
\usepackage{amsmath,amssymb,amsfonts}
\usepackage{algorithmic}
\usepackage{graphicx}
\usepackage{textcomp}
\usepackage{xcolor}
\def\BibTeX{{\rm B\kern-.05em{\sc i\kern-.025em b}\kern-.08em
    T\kern-.1667em\lower.7ex\hbox{E}\kern-.125emX}}
\begin{document}

\title{FMiFood: Multi-modal Contrastive Learning for Food Image Classification}


\author{
\IEEEauthorblockN{Xinyue Pan, \, Jiangpeng He, \, Fengqing Zhu}
\IEEEauthorblockA{
Elmore Family School of Electrical and Computer Engineering, \\
Purdue University, West Lafayette, IN, U.S.A. \\
\{pan161, he416, zhu0\}@purdue.edu
}}

\maketitle

\begin{abstract}
Food image classification is the fundamental step in image-based dietary assessment, which aims to estimate participants' nutrient intake from eating occasion images. A common challenge of food images is the intra-class diversity and inter-class similarity, which can significantly hinder classification performance. To address this issue, we introduce a novel multi-modal contrastive learning framework called FMiFood, which learns more discriminative features by integrating additional contextual information, such as food category text descriptions, to enhance classification accuracy. Specifically, we propose a flexible matching technique that improves the similarity matching between text and image embeddings to focus on multiple key information. Furthermore, we incorporate the classification objectives into the framework and explore the use of GPT-4 to enrich the text descriptions and provide more detailed context.  Our method demonstrates improved performance on both the UPMC-101 and VFN datasets compared to existing methods.

\vspace{-2mm}

\end{abstract}

\section{Introduction}
\vspace{-1mm}
Image-based dietary assessment involves measuring nutrient intake from food images captured during eating occasions and has been deployed in mobile and wearable technologies \cite{shao2021_ibdasystem, hamid2017, luke2015}. It also brings practical values to the nutrition science and healthcare fields, where assessment of a person's dietary intake plays an important role in the overall health of the individual \cite{reedy2014}.

\begin{figure}[t]
    \centering
    \includegraphics[width=0.8\linewidth]{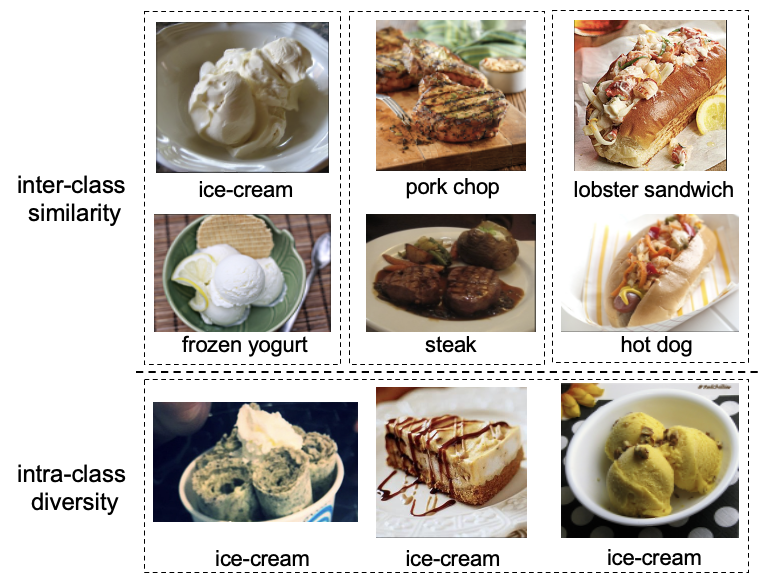}
    \caption{Examples of inter-class similarity and intra-class diversity.}
    \label{fig:issue}
\vspace{-5mm}
\end{figure}

The first step in image-based dietary assessment is food image classification, which has been extensively studied \cite{mao2020, pan2024_asilomar, Mao2021ImprovingDA, pan2023_madima,liu2024,min2023,jiang2020}. However, existing methods often encounter challenges of inter-class similarity and intra-class diversity, as illustrated in Fig. \ref{fig:issue}. To address these issues, contextual information can be leveraged to provide additional information that enhance the model's ability to learn more discriminative features from food images. In this paper, contextual information refers to food category text description, which is easy to obtain. 

Multi-modal contrastive learning (MMCL) aims to enhance classification by leveraging intrinsic relationships between different types of data, such as images and text. The core idea behind MMCL is to learn a joint embedding space where features from matched pairs (e.g., an image and its corresponding text description) are aligned, while the feature representations of unmatched pairs are pushed apart. This contrastive learning process encourages the model to capture the shared semantic information between the modalities and learn more discriminative features. MMCL is particularly effective for capturing the rich and diverse information present in multi-modal data because it simultaneously exploits the complementary strengths of visual and textual modalities. Images provide detailed visual context, capturing the appearance, texture, and spatial relationships of objects. On the other hand, text offers semantic and descriptive information, conveying the meaning, attributes, and relationships between concepts. By leveraging both modalities, MMCL enables the model to learn a more comprehensive and robust representation of the data.



However, most MMCL models \cite{radford2021, jia2021scaling, mu2021slip, yang2022} primarily focus on learning global relationships between image-text pairs, thus often struggle to handle noisy or irrelevant information in the text that is not directly related to the image. This limitation is particularly prevalent in the domain of food image classification, where images often contain complex components with intricate relationships. Even minor variations in the ingredients can lead to a change in the food category.
For example, as illustrated in fig \ref{fig:issue}, a small alteration in the ingredients of a lobster sandwich can result in its reclassification as a hot dog. Consequently, there is a need for a model capable of exploring the fine-grained relationships between image-text pairs to address these challenges.

To address these challenges, we propose the \textbf{F}lexible \textbf{M}atching for \textbf{i}mage Classification on \textbf{Food} images(FMiFood) model, which introduces a flexible matching mechanism that allows an image patch to match multiple text tokens or none, as appropriate. This flexibility enables our model to better capture the complexity and fine-grained details of food images, where ingredients may have intricate relationships. Inspired by~\cite{yang2022, wei2023}, we also combine the contrastive learning objective with a separate branch for the image classification objective, utilizing both the soft cross-entropy loss and the hard cross-entropy loss, respectively. To further enhance the richness and informativeness of the textual descriptions, we leverage the GPT-4 model~\cite{openai2023gpt4}, which possesses extensive knowledge about various food categories. By generating detailed and semantically meaningful descriptions for each label using GPT-4, our model can better understand and disambiguate the subtle differences between food images, leading to improved classification performance.


The main contributions of our work can be summarized as:
\begin{itemize}
    \item We propose the FMiFood model, specifically designed for the food image classification task, which utilizes soft cross-entropy loss in contrastive learning and includes a separate branch for the image classification objective.
    \item We introduce flexible matching in the context of contrastive learning to allow multiple or no text tokens matched to an image patch and vice versa as a new way to compute similarity scores between image-text pairs.
    \item We leverage the GPT-4 model to generate detailed text descriptions for each food category, enriching the textual information available for training.
\end{itemize}
\vspace{-1mm}

\section{Related work}
\vspace{-1mm}
\subsection{Food Image Classification}
Many contributions have been made to food image classification. R. Mao \textit{et al.} proposed visual hierarchy and multi-task learning for food classification\cite{mao2020,Mao2021ImprovingDA}. M. Wei \textit{et al.} introduced ISIA-500\cite{Min-ISIA-500-MM2020} and Food-2K\cite{min2023} datasets, and enhanced local feature extraction\cite{min2023}. S. Jiang \textit{et al.} aggregated multi-level features for improved classification\cite{jiang2020}. These approaches address intra-class diversity and inter-class similarity issues. J. He \textit{et al.} focused on food continual learning~\cite{ILIO, he2021_iccvw} and imbalanced classification\cite{he2023singlestage, he2023_long_tailed} to simulate real world food data distribution.


\vspace{-2mm}
\subsection{Multi-Modal Contrastive Learning}
Multi-modal contrastive learning has gained attention in image classification due to its significant improvements in tasks such as image-text retrieval and image classification compared to using image data alone. One of the widely recognized models in this field is CLIP \cite{radford2021}, which aligns features of matched image-text pairs and separates unmatched pairs. Other models have introduced various improvements based on CLIP, such as ALIGN, UNiCL, SLIP and DeCLIP \cite{jia2021scaling, yang2022,mu2021slip, li2022supervision}. However, these models typically focus on global relationships between image-text pairs and do not explore fine-grained relationships, which are crucial for food image classification.

FILIP \cite{yao2022} is designed to capture fine-grained relationships between images and text by focusing on alignments between image patches and text tokens. FILIP achieves this by basing similarity scores on the mean max token-wise similarity between image patches and text tokens. However, FILIP is not specifically designed for image classification and can be generalized to other downstream tasks, such as image-text retrieval. FILIP assumes that each image matches only one text and vice versa within a single batch, which does not account for scenarios where multiple texts can match a single image and multiple images can belong to the same category. Additionally, FILIP assumes each image patch matches only one text token, which is also not always the case. FILIP's classification performance is further limited by a lack of detailed text descriptions for each text label. Therefore, while FILIP partially addresses issues of intra-class diversity and inter-class similarity, there is still space for improvement.

iCLIP \cite{wei2023} and UniCL \cite{yang2022} are MMCL models specifically designed for image classification. iCLIP enhances label explanations using the WordNet dictionary, a large lexical database of English that groups words into sets of synonyms called synsets. It also adapts the CLIP model for image classification by comparing images in a batch with all label descriptions to compute the cross-entropy loss. However, this approach is not suitable for food categories since most food categories lack corresponding explanations in the WordNet \cite{miller-1994-wordnet} dictionary. Additionally, iCLIP cannot handle multiple positive pairs associated with one image or text label in the contrastive learning component. Although UniCL addresses this issue by applying soft cross-entropy loss during the training phase, it also has limitations as it cannot consider all possible negative samples (labels) in a single batch. These limitations highlight the need for improved methods to address these challenges in MMCL for image classification.
\vspace{-1mm}

\section{Method}

\begin{figure*}[t]
    \centering
    \includegraphics[width=1.01\linewidth]{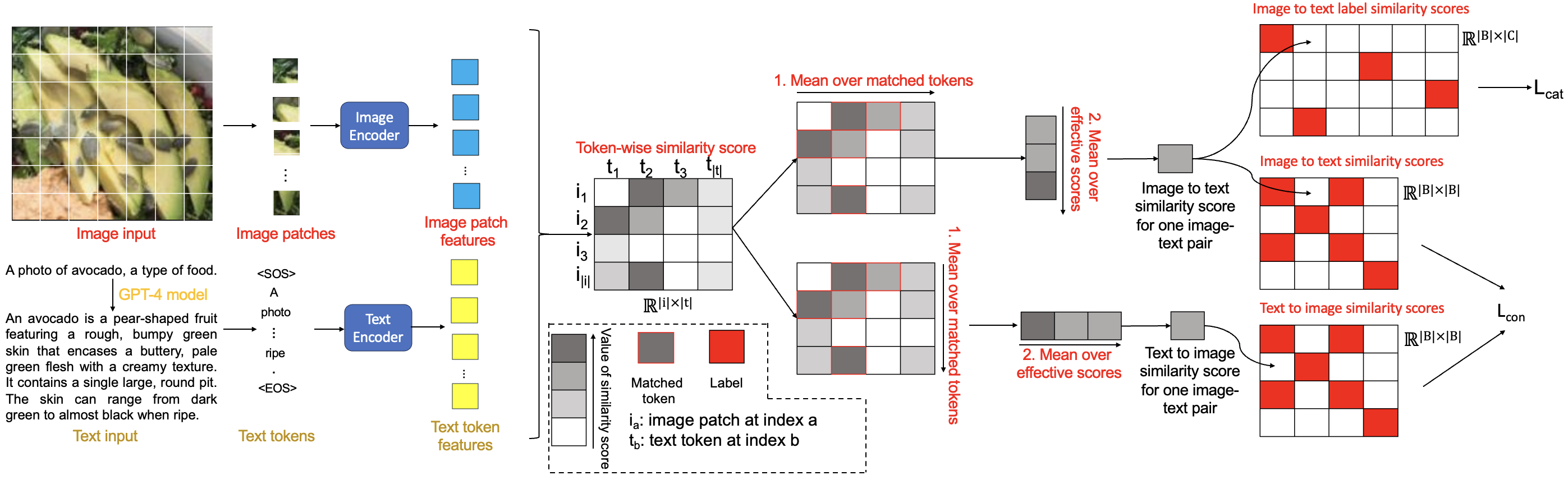}
    \caption{Overview of our FMiFood model: $|B|$ denotes the batch size and $|C|$ represents the number of text labels in the dataset. Images and text descriptions are fed into the image encoders and text encoders of the FMiFood model to extract image patch and text token or label token features. The similarity score between image-text pairs is computed based on the flexible matching technique to learn with both the contrastive loss and the categorical loss.}
    \label{fig:method}
    \vspace{-3mm}
\end{figure*}

\vspace{-1mm}
In this section, we introduce our FMiFood framework as shown in Fig. \ref{fig:method}. The model computes similarity scores between image-text pairs using a flexible matching technique and incorporates an image classification learning objective, which are illustrated in 
Section \ref{sec:fm} and Section \ref{sec:filip_img}, respectively. We also augment text descriptions using the GPT-4 model to explain food categories as detailed in Section \ref{sec:text_des}. 

\begin{figure}[t]
    \centering
    \includegraphics[width=0.75\linewidth]{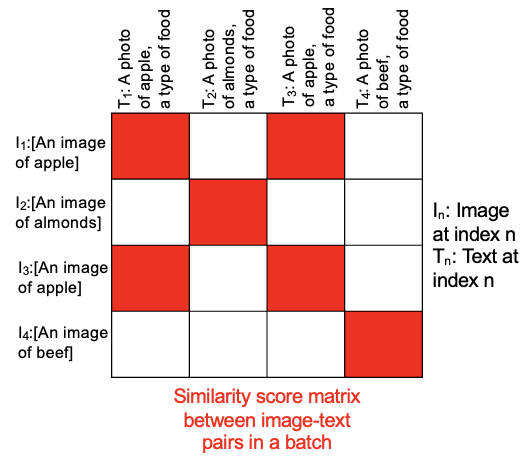}
    \caption{Issue with contrastive learning on current multi-modal contrastive learning model: In a single batch, we cannot assume one image is only matched to one text in contrastive learning under image classification task.}
    \label{fig:method_sce}
\vspace{-3mm}
\end{figure}
\vspace{-2mm}
\subsection{Flexible Matching Between Image Patches and Text Tokens}
\label{sec:fm}
To enhance the alignment between image patches and text tokens, we use a flexible matching strategy. This approach allows an image patch to match multiple text tokens or none at all, and vice versa. Instead of considering only the maximum similarity score between an image patch and all text tokens, we average all similarity scores above a certain threshold while ignoring those below a lower threshold. This ensures that all relevant matches are included, capturing the complex relationships between image patches and text tokens more accurately.


Let $I$ be the set of image patches and $T$ be the set of text tokens. Let $s(i, t)$ be the similarity score between image patch $i \in I$ and text token $t \in T$. The effective number of image patches and text tokens taken into account for similarity score computation are denoted as $|I|$ and $|T|$, respectively. The similarity score between image patches and text tokens is computed as follows. The similarity score from image to text, $\text{sim}(I, T)$, is given as
\begin{equation}
\text{sim}(I, T) = \frac{1}{|I|} \sum_{i \in I} s_i
\vspace{-1mm}
\end{equation}
where
\begin{equation}
s_i =
\begin{cases}
    \frac{1}{|T_{d}^i|} \sum_{t \in T_{d}^i} s(i, t) & \text{if } \max(s(i, t)) > d, \\
    \max_{t \in T_{[c,d]}^i} s(i, t) & \text{if } c \leq \max(s(i, t)) \leq d, \\
    0 & \text{if } \max(s(i, t)) < c,
\end{cases}
\vspace{-1mm}
\end{equation}
and
\begin{equation}
T_{d}^i = \{t \in T \mid s(i, t) > d\},
\vspace{-1mm}
\end{equation}
\begin{equation}
T_{[c,d]}^i = \{t \in T \mid c \leq s(i, t) \leq d\}.
\vspace{-1mm}
\end{equation}

Similarly, the similarity score from text to image, $\text{sim}(T, I)$, can be defined analogously:
\begin{equation}
\text{sim}(T, I) = \frac{1}{|T|} \sum_{t \in T} s_t
\vspace{-1mm}
\end{equation}
where
\begin{equation}
s_t =
\begin{cases}
    \frac{1}{|I_{d}^t|} \sum_{i \in I_{d}^t} s(t, i) & \text{if } \max(s(t, i)) > d, \\
    \max_{i \in I_{[c,d]}^t} s(t, i) & \text{if } c \leq \max(s(t, i)) \leq d, \\
    0 & \text{if } \max(s(t, i)) < c,
\end{cases}
\vspace{-1mm}
\end{equation}
and
\begin{equation}
I_{d}^t = \{i \in I \mid s(t, i) > d\},
\vspace{-1mm}
\end{equation}
\begin{equation}
I_{[c,d]}^t = \{i \in I \mid c \leq s(t, i) \leq d\}.
\vspace{-1mm}
\end{equation}
where $s(i, t)$ is the cosine similarity score between image patch $i$ and text token $t$ and $-1 \leq s(i, t) \leq 1$. $\max(s(i, t))$ is the maximum similarity score between image patch $i$ and all text tokens $t \in T$, $\max(s(t, i))$ is the maximum similarity score between text token $t$ and all image patches $i \in I$. $T_{d}^i$ and $T_{[c,d]}^i$ are the sets of text tokens with similarity scores above d and between c and d with image patch $i$, respectively. $I_{d}^t$ and $I_{[c,d]}^t$ are the sets of image patches with similarity scores above d and between c and d with text token $t$, respectively.
\vspace{-1mm}
\subsection{Image Classification Learning Objective}
\label{sec:filip_img}

In the image classification task, text information typically comes from category labels. To effectively manage the scenario where multiple images can match the same text label within a single batch, as shown in fig. \ref{fig:method_sce}, we apply a soft cross-entropy loss in FMiFood's contrastive learning, inspired by the UniCL model \cite{yang2022}. The image-to-text contrastive loss $\mathcal{L}^I_k$ for $x^I_k$ is given by:
\begin{equation}
\mathcal{L}^I_k \left( x^I_k, \{x^T_j\}^b_{j=1} \right) = - \frac{1}{b} \sum_{j=1}^b q^I_{k,j} \log \left( \frac{\exp(s^I_{k,j})}{\sum_{l=1}^b \exp(s^I_{k,l})} \right)
\vspace{-1mm}
\end{equation}
where $s^I_{k,j}$ denotes the similarity of the $k$-th image to the $j$-th text, and $q^I_{k,j}$ represents the ground truth indicator for whether the $k$-th image and the $j$-th text are a positive pair.

The text-to-image contrastive loss for $x^T_k$ is:
\begin{equation}
\mathcal{L}^T_k \left( x^T_k, \{x^I_j\}^b_{j=1} \right) = - \frac{1}{b} \sum_{j=1}^b q^T_{k,j} \log \left( \frac{\exp(s^T_{j,k})}{\sum_{l=1}^b \exp(s^T_{l,k})} \right)
\vspace{-1mm}
\end{equation}
where $b$ is the batch size ,and $s^T_{j,k}$ denotes the similarity of the $j$-th text to the $k$-th image, and $q^T_{k,j}$ represents the ground truth indicator for whether the $k$-th text and the $j$-th image are a positive pair.

The total contrastive loss is then:
\begin{equation}
\mathcal{L_{\text{con}}} = \frac{1}{2} \sum_{k=1}^b \left( \mathcal{L}^I_k + \mathcal{L}^T_k \right)
\vspace{-1mm}
\end{equation}

However, contrastive learning alone may not see all negative classes in a batch. To address this, we incorporate an additional image classification objective. By computing the cosine similarity between image and text labels and applying cross-entropy loss, we directly optimize the model for image classification. The predicted probability for the $k$-th image belonging to the $j$-th category is given by:
\begin{equation}
p(y_k = j \mid x^I_k) = \frac{\exp(s^I_{k,j})}{\sum_{l=1}^C \exp(s^I_{k,l})}
\vspace{-1mm}
\end{equation}
where $C$ is the total number of categories. The categorical cross-entropy loss for a batch of $b$ images is:
\begin{equation}
\mathcal{L}_{\text{cat}} = \frac{1}{b} \sum_{k=1}^b \mathcal{L}_{\text{cat}, k} = - \frac{1}{b} \sum_{k=1}^b \log \left( \frac{\exp(s^I_{k, y_k})}{\sum_{l=1}^C \exp(s^I_{k,l})} \right)
\vspace{-1mm}
\end{equation}
where $s^I_{k,y_k}$ denotes the similarity of the $k$-th image to its true category label $y_k$.

Finally, the total loss combines both losses:

\begin{equation}
\mathcal{L}_{\text{total}} = m*\mathcal{L}_{\text{con}} + n*\mathcal{L}_{\text{cat}}
\vspace{-1mm}
\end{equation}
where $m$ and $n$ are hyperparameters associated with each loss and can be tuned.

\vspace{-1mm}
\subsection{Generation of Text Descriptions Using GPT-4}
\label{sec:text_des}
\vspace{-1mm}
Typical multi-modal contrastive learning models use standard text templates like "A photo of [category], a type of food." These text templates lack detailed information, which can limit the model's ability to learn nuanced features of different categories. To address this, we enhance text descriptions by leveraging the GPT-4 model \cite{openai2023gpt4}, which is pretrained on a vast amount of real-world data. By prompting GPT-4 with "Describe the appearance of [category] in less than 50 words," we generate detailed descriptions that provide richer context for the model to learn. GPT-4 can generate more specific descriptions that can offer more informative visual cues, helping the model to better distinguish between similar categories. Incorporating these enriched descriptions into the training data improves the alignment between image and text features, ultimately leading to enhanced classification performance.
\vspace{-1mm}

\section{Experiments}

\vspace{-0.5mm}
We evaluate our FMiFood model using the average classification accuracy on two datasets to conduct ablation studies and compare the performance to existing works.

\subsection{Datasets}

\textbf{UPMC-Food101}\cite{gallo2020}: This dataset contains 101 food categories and 790 - 956 images per category with food recipe descriptions extracted from the Internet for each image, which include noisy information associated with each image. However, considering the generalization of our method and the fact that most food datasets do not have such descriptions, we do not use the text descriptions in this dataset for our experiments. The training and testing sets are preset in this dataset.

\textbf{VFN}\cite{mao2020}: This dataset contains 74 food categories with around 14K images. Each food category represents a frequently consumed food term selected from the What We Eat In America (WWEIA) database\cite{eicher2017}. All food images are online images uploaded by real users. We randomly split the dataset into 80:20 as training and testing sets.
\vspace{-2mm}
\subsection{Experiment Setup}
We utilize the Vision Transformer (ViT)\cite{dosovitskiy2021image} with a patch size of 32 as the baseline model for uni-modal learning. For multi-modal contrastive learning, we use CLIP\cite{radford2021}, FILIP\cite{yao2022}, UNiCL\cite{yang2022} and iCLIP\cite{wei2023} as baseline models.

\begin{itemize}
    \item \textbf{ViT}\cite{dosovitskiy2021image}: A Vision Transformer model with a patch size of 32 that utilizes only image information for training.
    \item \textbf{CLIP}\cite{radford2021}: Conducts contrastive learning on image-text pairs to make matched pairs as close as possible while separating unmatched pairs based on global features. The text input used is "A photo of [category], a type of food."
    \item \textbf{FILIP}\cite{yao2022}: Conducts contrastive learning on image-text pairs like CLIP but based on fine-grained relationships using image patch features and text token features. The text input used is "A photo of [category], a type of food."
    \item \textbf{UniCL}\cite{yang2022}: A model that considers multiple texts matched to an image within a batch during the contrastive learning process. The text input used is "A photo of [category], a type of food."
    \item \textbf{iCLIP}\cite{wei2023}: A CLIP model applied to image classification tasks with augmented text descriptions from the WordNet dictionary. The text input consists of the text label along with its corresponding explanation from the WordNet dictionary.
\end{itemize}

We also conduct ablation studies on related work and our proposed methods. We use two types of text descriptions: the standard one and the GPT-4 augmented one. The standard text description is ``A photo of [category], a type of food.'' The GPT-4 augmented description is ``A photo of [category], a type of food. [GPT-4 generated descriptions].'' Our proposed method includes flexible matching (FM) as detailed in Section \ref{sec:fm} and the image classification learning objective(IC) detailed in Section \ref{sec:filip_img}. For flexible matching, we emperically set the lower threshold $c=0$ and upper threshold $d=0.85$. For image classification learning objective(IC), we set hyper-parameter $m=0.75$ and $n=0.25$.

The image encoder used in the multi-modal contrastive learning models is a Vision Transformer with a patch size of 32, and the text encoder is a masked self-attention Transformer from CLIP. We train the baseline models and the proposed method on the UPMC-Food101 \cite{gallo2020} and VFN \cite{mao2020} food datasets. The batch size is set to 128, and we use the AdamW optimizer with an initial learning rate of $1e-5$ with a cosine annealing scheduler \cite{loshchilov2017sgdr}. The model is trained for 40 epochs across different methods.

\begin{table}[t]
\begin{center}
\caption{Average classification accuracy in percentage on two datasets for different methods} \label{tab:result_all}
\begin{tabular}{c|c|c|c}
\hline
\makecell{Method} & \makecell{UPMC-Food101} & \makecell{VFN} & \makecell{Average}\\
\hline
ViT\cite{dosovitskiy2021image} & 69.17 & 75.36 & 72.27\\
CLIP\cite{radford2021} & 74.87 & 79.78 & 77.32\\
FILIP\cite{yao2022} & 75.10 & 79.51 & 77.31\\
UniCL\cite{yang2022} & 75.53 & 80.24 & 77.89\\
iCLIP\cite{wei2023} & 75.72 & 80.70 & 78.21\\
\hline
\textbf{FMiFood} & \textbf{76.22} & \textbf{81.69} & \textbf{78.96}\\
\hline
\end{tabular}
\end{center}
\vspace{-5mm}
\end{table}

\begin{table}[t]
\begin{center}
\caption{Average classification accuracy in percentage on UPMC-Food101 dataset for ablation studies on our proposed method} \label{tab:result_upmc}
\begin{tabular}{c|c|c|c|c}
\hline
\makecell{Method} & \makecell{FM} & \makecell{IC} & \makecell{Accuracy for \\ Standard \\ text description} & \makecell{Accuracy for \\GPT-4 augmented \\ text description}\\
\hline
CLIP\cite{radford2021} & - & - & 74.87 & 73.70 \\
FILIP\cite{yao2022} & - & - & 75.10 & 75.59\\
FMiFood & Yes & No & 75.61 & 75.76\\
FMiFood & No & Yes & 75.75 & 75.92\\
\textbf{FMiFood} & Yes & Yes & \textbf{76.06} & \textbf{76.22}\\
\hline
\end{tabular}
\end{center}
\vspace{-5mm}
\end{table}

\begin{table}[t]
\begin{center}
\caption{Average classification accuracy in percentage on VFN dataset for ablation studies on our proposed method} \label{tab:result_vfn}
\begin{tabular}{c|c|c|c|c}
\hline
\makecell{Method} & \makecell{FM} & \makecell{IC} & \makecell{Accuracy for \\ Standard \\ text description} & \makecell{Accuracy for \\GPT-4 augmented \\ text description}\\
\hline
CLIP\cite{radford2021} & - & - & 79.78 & 80.80 \\
FILIP\cite{yao2022} & - & - & 79.51 & 80.60\\
FMiFood & Yes & No & 80.34 & 80.90\\
FMiFood & No & Yes & 80.86 & 81.32\\
\textbf{FMiFood} & Yes & Yes & \textbf{81.26} & \textbf{81.69}\\
\hline
\end{tabular}
\end{center}
\vspace{-5mm}
\end{table}

\begin{figure}[t]
    \centering
    \includegraphics[width=1\linewidth]{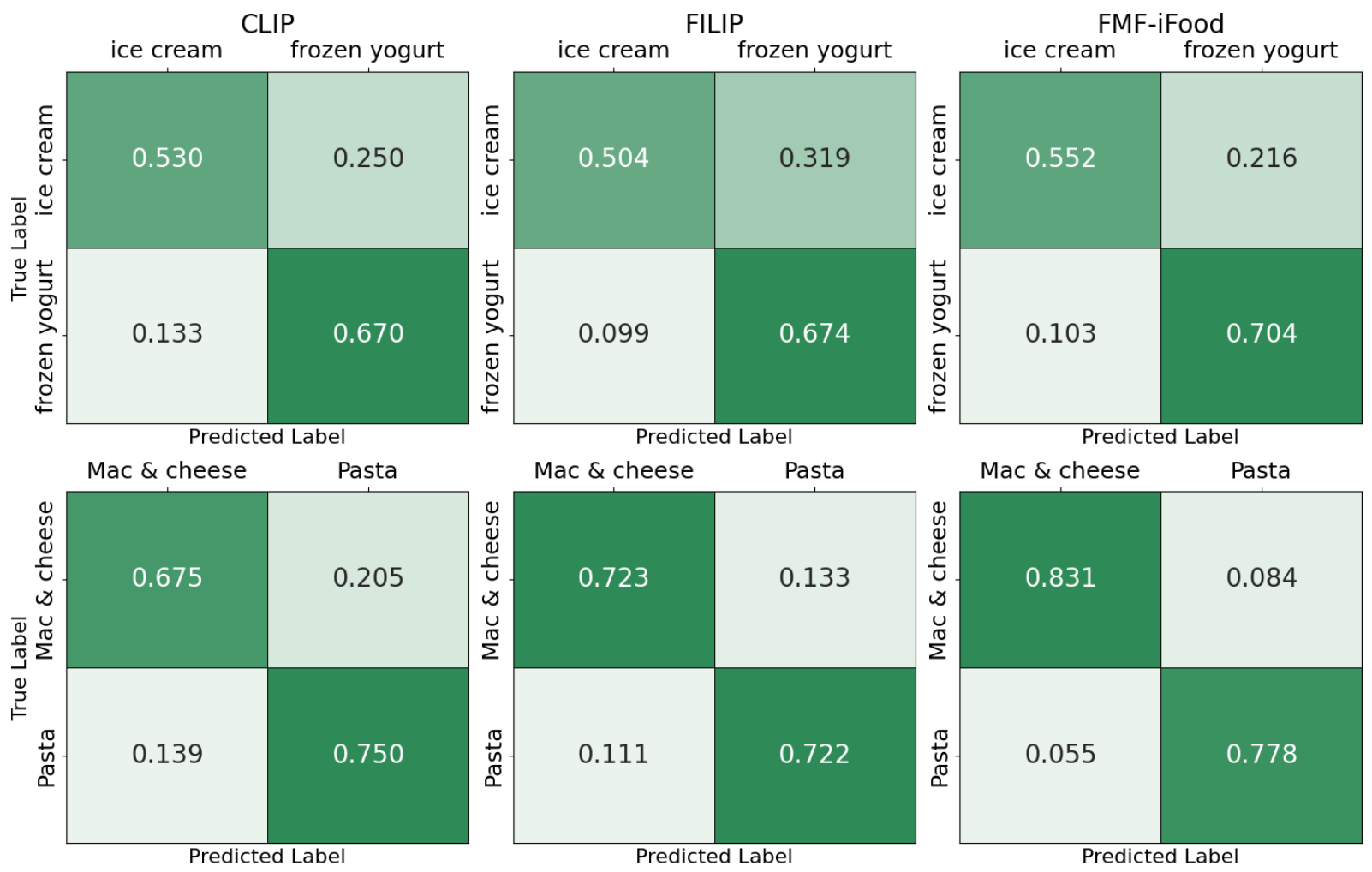}
    \caption{Partial confusion matrix for selected categories from the UPMC-Food101 and VFN datasets for different methods}
    \label{fig:upmc_confus}
\vspace{-5mm}
\end{figure}

\begin{figure}[t]
    \centering
    \includegraphics[width=1\linewidth]{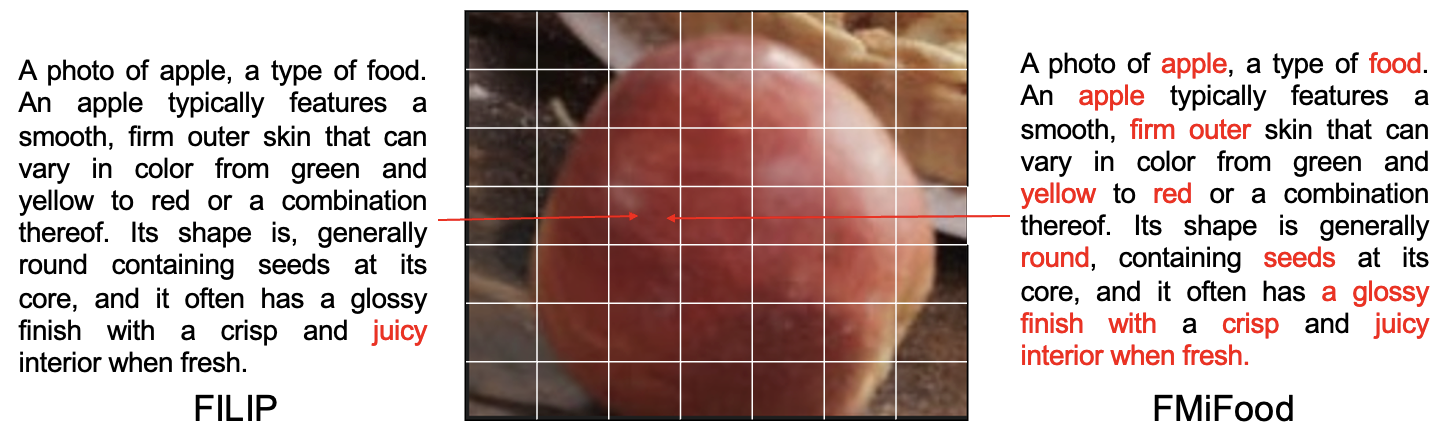}
    \vspace{-8mm}
    \caption{Qualitative result of comparison between FILIP and FMiFood. The words in red are the text tokens that are matched to the image patch}
    \label{fig:fm_result}
    \vspace{-3mm}
\end{figure}
\subsection{Quantitative Results and Ablation Studies}

Tables \ref{tab:result_all}, \ref{tab:result_upmc}, and \ref{tab:result_vfn} show the quantitative results in terms of average classification accuracy for different methods and ablation studies on the UPMC-Food101 and VFN datasets. Our proposed FMiFood model outperforms all related works in terms of average classification accuracy. 

From the results on the UPMC-Food101 dataset, we observe that augmenting text descriptions with the GPT-4 model does not improve CLIP's classification performance. This is because the text descriptions generated by the GPT-4 model on the UPMC-Food101 dataset are not directly related to the image, making CLIP vulnerable to noisy information in the text. However, for FILIP and FMiFood models, there is a small improvement after augmenting text descriptions with the GPT-4 model. This is because these models can filter out irrelevant information through the matching process between image patches and text tokens. Specifically, FMiFood can effectively filter out image patches that do not correspond to any relevant text tokens and vice versa. In all scenarios, incorporating the image classification learning objective significantly improves classification accuracy. This is because the model specifically focuses on optimizing for image classification, ensuring that the features learned are highly relevant to the task.

For the VFN dataset, we observe that CLIP achieves much better classification performance after augmenting text descriptions with the GPT-4 model. This is because the generated text descriptions are related to the image content. This improvement is also seen in FILIP and FMiFood models, but the improvement is smaller. This is because FILIP and FMiFood models do not consider all important information when matching image patches and text tokens, but FMiFood can mitigate this issue.

Fig \ref{fig:upmc_confus} presents the partial confusion matrices for selected categories from the UPMC-Food101 and VFN datasets. On the UPMC-Food101 dataset, it can be seen that the `ice-cream' category is often misclassified as `frozen yogurt' by the CLIP and FILIP models due to the visual similarity between these two categories. However, the FMiFood model demonstrates better performance in distinguishing between these categories. A similar phenomenon is observed for the `Macaroni and noodles with cheese' category, which is often confused with `pasta mixed dishes' in the CLIP and FILIP models. The improved performance of FMiFood can be attributed to its ability to leverage richer text descriptions for each food category, thereby enhancing its discriminatory power between visually similar food items.

\vspace{-1mm}
\subsection{Qualitative Results for Flexible Matching in FMiFood}

Fig \ref{fig:fm_result} shows qualitative results of matching between one image patch and all text tokens on one example food image with or without the flexible matching method. From FILIP's results, we see that one image patch can only be matched to one text token, losing much information in the text. However, for FMiFood, one image patch can be matched to multiple text tokens, which includes rich information and captures key information from the text. There are some matched text tokens not directly related to the image, introducing noisy information during the learning process. This is also why FMiFood does not contribute much performance gain in addition to the FILIP model. Therefore, future work will focus on investigating how to filter out unrelated information in the matched tokens.
\vspace{-1mm}

\section{Conclusion and future work}


\vspace{-0.5mm}
In this paper, we apply a multi-modal contrastive learning model to the task of food image classification. To address the issues of intra-class diversity and inter-class similarity, we propose the FMiFood model, which incorporates fine-grained learning. To learn richer text information and filter out irrelevant tokens, we introduce flexible matching between image patches and text tokens, extending beyond the capabilities of the FILIP model. Additionally, to specifically target the image classification learning objective, we apply soft cross-entropy loss in the contrastive learning component of the FMiFood model. This allows the model to consider cases where multiple images in a batch match a single text label. We also incorporate categorical loss to facilitate direct classification during the learning process.  Our experimental results demonstrate good improvements in classification accuracy on two food datasets. In the future, we aim to enhance the flexible matching method in the FMiFood model to ensure that the matched tokens do not include noisy information, further improving the model’s robustness and performance.
\vspace{-0.5mm}

\bibliographystyle{IEEEtran}
\small \bibliography{main}

\end{document}